\documentclass[runningheads,a4paper]{llncs}

\usepackage{amssymb}
\usepackage{graphicx}
\usepackage{subfigure}
\usepackage{booktabs}
\usepackage{multirow}
\usepackage{color}

\usepackage{url}
\urldef{\mailsb}\path|{eduardo.aguilar,marc.bolanos,petia.ivanova}@ub.edu | 
\newcommand{\keywords}[1]{\par\addvspace\baselineskip
\noindent\keywordname\enspace\ignorespaces#1}

\begin{document}

\mainmatter  

\title{Food Recognition using Fusion of Classifiers based on CNNs}

\titlerunning{Food Recognition using Fusion of Classifiers based on CNNs}

%
%
\author{Eduardo Aguilar
%
\and Marc Bola\~nos
\and Petia Radeva
}
%

\institute{Universitat de Barcelona and Computer Vision Center, Spain
\\
\mailsb\\
}
%
%

\toctitle{Lecture Notes in Computer Science}
\tocauthor{Authors' Instructions}
\maketitle

\begin{abstract}
With the arrival of convolutional neural networks, the complex problem of food recognition has experienced an important improvement in recent years. The best results have been obtained using methods based on very deep convolutional ceural cetworks, which show that the deeper the model,the better the classification accuracy will be obtain. However, very deep neural networks may suffer from the overfitting problem. In this paper, we propose a combination of multiple classifiers based on different convolutional models that complement each other and thus, achieve an improvement in performance. The evaluation of our approach is done on two public datasets: Food-101 as a dataset with a wide variety of fine-grained dishes, and Food-11 as a dataset of high-level food categories, where our approach outperforms the independent CNN models. 


\end{abstract}
\section{Introduction}

In the field of computer vision, food recognition has caused a lot of interest for researchers considering its applicability in solutions that improve people's nutrition and hence, their lifestyle \cite{who2014}. In relation to the healthy diet, traditional strategies for analyzing food consumption are based on self-reporting and manual quantification \cite{shim2014}. Hence, the information used to be inaccurate and incomplete \cite{rumpler2008}. Having an automatic monitoring system and being able to control the food consumption is of vital importance, especially for the treatment of individuals who have eating disorders, want to improve their diet or reduce their weight.

Food recognition is a key element within a food consumption monitoring system. Originally, it has been  approached by using traditional approaches \cite{bossard2014,liu2016}, which extracted ad-hoc image features by means of algorithms based mainly on color, texture and shape. More recently, other approaches focused on using Deep Learning techniques \cite{liu2016,yanai2015,martinel2016,hassannejad2016}. In these works, feature extraction algorithms are not hand-crafted and additionally, the models automatically learn the best way to discriminate the different classes to be classified. As for the results obtained, there is a great difference (more than 30\%) between the best method based on hand-crafted features compared to newer methods based on Deep Learning, where the best results have been obtained with Convolutional Neural Networks (CNN) architectures that used inception modules \cite{hassannejad2016} or residual networks \cite{martinel2016}.

\begin{figure}[t!]
\centering
\includegraphics[width=1\textwidth]{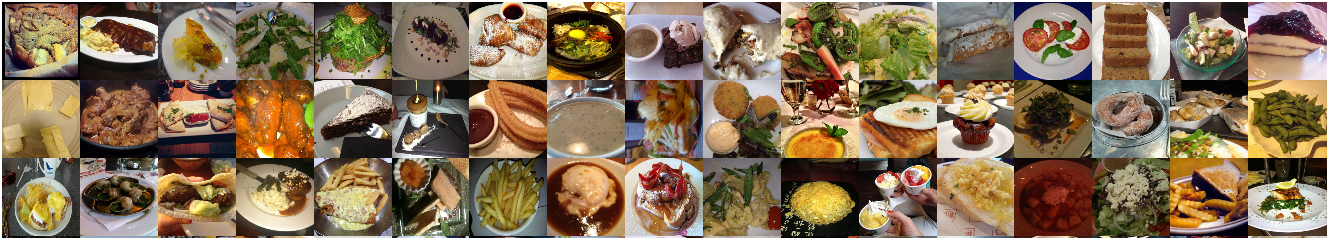}
\caption{\small Example images of Food-101 dataset. Each image represents a dish class.}
\label{fig:fig3}
\end{figure}

Food recognition can be considered as a special case of object recognition, being the most active topic in computer vision lately. The specific part is that dish classes have a much higher inter-class similarity and intra-class variation than usual ImageNet objects (cars, animals, rigid objects, etc.) (see Fig.\ref{fig:fig3}). If we analyze the last accuracy increase in the ImageNet Large Scale Visual Recognition Challenge (ILSVRC) \cite{russakovsky2015}, it has been improving thanks to the depth increase of CNN models \cite{krizhevsky2012,zeiler2014,szegedy2015,he2016} and also thanks to the use of CNNs fusion strategies\cite{zeiler2014,he2016}. The main problem of CNNs is the need of large datasets to avoid overfitting the network as well as the need of high computational power for training them. 

Considering the use of different networks trained on the same data, one can observe that patterns misclassified by the different models would not necessarily overlap \cite{kittler1998}. This suggests that different classifiers could potentially offer complementary information that can be used to improve the final performance \cite{kittler1998}. An option to combine the outputs of different classifiers was proposed in \cite{kuncheva2001}, where the authors used what they call a decision templates scheme instead of simple aggregation operators such as the product or average. As they showed, this scheme maintains a good performance using different training set sizes and is also less sensitive to particular datasets compared to the other schemes.

In this article, we introduce the fusion concept to the CNN framework, with the purpose of demonstrating that the combination of the classifiers' output, by using a decision template scheme, allows to improve the performance on the food recognition problem.
Our contributions are the following: 1) we propose the first food recognition algorithm that fuses the output of different CNN models, 2) we show that our CNNs Fusion approach has better performance compared to the use of CNN models separately, and 3) we demonstrate that our CNNs Fusion approach keeps a high performance independently of the target (dishes, family of dishes) and dataset (validating it on 2 public datasets). 

The organization of the article is as follows. In section \ref{sec:methodology}, we present the CNNs Fusion methodology. In section \ref{sec:experiments}, we present the datasets, the experimental setup and discuss the results. Finally, in section \ref{sec:conclusions}, we describe the conclusions.

\begin{figure}
\centering
\includegraphics[width=1\textwidth]{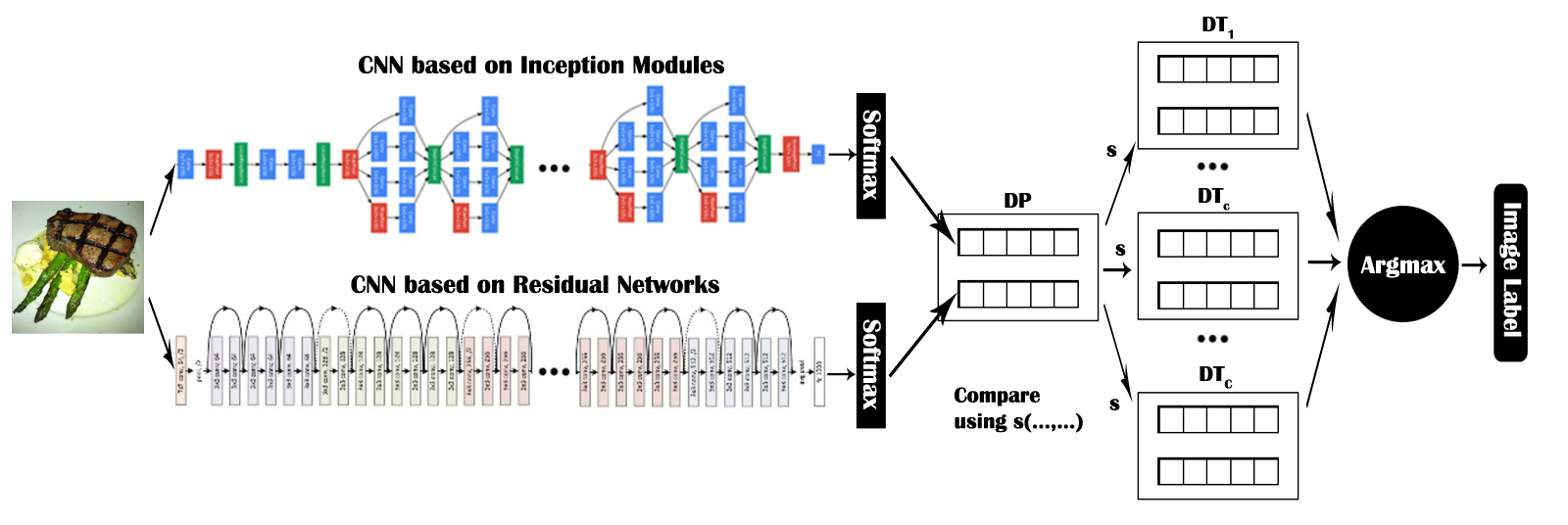}
\caption{\small General scheme of our CNNs Fusion approach.}
\label{fig:fig1}
\end{figure}

\section{Methodology}\label{sec:methodology}

In this section, we describe the CNN Fusion methodology (see Fig. \ref{fig:fig1}), which is composed of two main steps: training $K$ CNN models based on different architectures and fusing the CNN outputs using the decision templates scheme.

\subsection{Training of CNN models}\label{sec:training}

The first step in our methodology involves separately training two CNN models. We chose two different kind of models winners of the ILSVRC in the object recognition task. Both models won or are based on the winner of the challenges made in 2014 and 2015 proposing novel architectures: the first based its design in "inception models" and the second in "residual networks". First, each model was pre-trained on the ILSVRC data. Later, all layers were fine-tuned by a certain number of epochs, selecting for each one the model that provides the best results in the validation set and that will be used in the fusion step.

\subsection{Decision templates for classifiers fusion}\label{sec:fusion}

Once we trained the models on the food dataset, we combined the softmax classifier outputs of each model using the Decision Template (DT) scheme \cite{kuncheva2001}.

Let us annotate the  output of the last layer of the $k$-th CNN model as $(\omega _{1,k},\ldots,\omega _{C,k})$, where $c=1,...,C$ is the number of classes and $k=1,...K$ is the index of the CNN model (in our case, K=2). Usually, the softmax function is applied, to obtain the probability value of model $k$ to classify image $x$  to a class $c$: 
$
p_{k,c}(x) = \frac{e^{\omega _{k,c}}}{\sum_{c=1}^{C} {e}^{\omega _{k,c}}}.
$
Let us consider the $k$-th decision vector $D_{k}$:

\[
D_{k}(x) = [p_{k,1}(x), p_{k,2}(x), ..., p_{k,C}(x)]
\]

{\bf Definition  \cite{kuncheva2001}:} A  {\bf Decision Profile,} DP for a given image $x$ is defined as:
\begin{equation}\label{eq:eq03}
DP(x) =
\left[ \begin{array}{cccc}
p_{1,1}(x) & p_{1,2}(x) & ... &  p_{1,C}(x) \\
 & \ldots &
\\
p_{K,1}(x) & p_{K,2}(x) & ... & p_{K,C}(x)
\end{array} \right]
\end{equation}

{\bf Definition  \cite{kuncheva2001}:} Given $N$ training images, a {\bf Decision Template} is defined as a set of matrices $DT=(DT^1,\ldots,DT^C)$, where the $c$-th element is obtained as the average of the decision profiles (\ref{eq:eq03}) on the training images of class $c$:  
\[
DT^{c} = \frac{\sum_{j =1}^N {DP(x_j)\times Ind(x_j,c)}}{\sum _{j=1}^N Ind(x_j,c)},
\]

where $Ind(x_j,c)$ is an indicator function with value 1 if
the training image $x_j$ has a crisp label $c$, and 0, otherwise \cite{kuncheva1995}. 

Finally, the resulting prediction for each image is determined considering the similarity $s(DP(x),DT^c(x))$  between the decision profile $DP(x)$ of the test image and the decision template of class $c,$ $c=1,\ldots,C$. 
Regarding the arguments of the similarity function $s(.,.)$ as fuzzy sets on some universal set with $K \times C$ elements, various fuzzy measures of similarity can be used.
We chose different measures  \cite{kuncheva2001}, namely 2 measures of similarity, 2 inclusion indices, a consistency measure and the Euclidean Distance. These measures are formally defined as: 

\[
S_{1}(DT^{c}, DP(x)) = \frac{\sum_{k=1}^K\sum_{i=1}^C \min(DT_{k,i}^c, DP_{k,i}(x))}{\sum_{k=1}^K\sum_{i=1}^C \max(DT_{k,i}^c, DP_{k,i}(x))},  
\]
\[
S_{2}(DT^{c}, DP(x)) = 1- \sup_u\{\left|DT_{k,i}^c-DP_{k,i}(x)\right| : c=1,\ldots,C , k=1,\ldots,K\},
\]

\[
I_{1}(DT^{c}, DP(x)) = \frac{\sum_{k=1}^K\sum_{i=1}^C \min(DT_{k,i}^c, DP_{k,i}(x))}{\sum_{k=1}^K\sum_{i=1}^C DT_{k,i}^c},
\]
\[
I_{2}(DT^{c}, DP(x)) = \inf_u\{\max(\overline{DT_{k,i}^c},DP_{k,i}(x)) : c=1,\ldots,C, k=1,\ldots,K\},
\]

\[
C(DT^{c}, DP(x)) = \sup_u\{\min(DT_{k,i}^c,DP_{k,i}(x)) :  c=1,\ldots,C, k=1,\ldots,K\},
\]
\[
N(DT^{c}, DP(x)) = 1 - \frac{\sum_{k=1}^K\sum_{i=1}^C (DT_{k,i}^c-DP_{k,i}(x))^2}{K \times C},  
\]

where $DT_{k,i}^c$ is the probability assigned to the class  $i$ by the classifier $k$ in the $DT^c$, $\overline{DT_{k,i}^c}$ is the complement of $DT_{k,i}^c$ calculated as $1-DT_{k,i}^c$, and $DP_{k,i}(x)$ is the probability assigned by the classifier $k$ to the class $i$ in the DP calculated for the image, $x$. The final label, $L$ is obtained as the class that maximizes the similarity, $s$ between $DP(x)$ and $DT^c$:
$L(x) = argmax _{c=1,\ldots,C}\{s(DT^c, DP(x))\}.$

\section{Experiments}\label{sec:experiments}

\subsection{Datasets} 

The data used to evaluate our approach are two very different images: Food-11 \cite{singla2016} and Food-101 \cite{bossard2014}, which are chosen in order to verify that the classifiers fusion provides good results regardless of the different properties of the target datasets, such as intra-class variability (the first one is composed of many dishes of the same general category, while the second one is composed of specific fine-grained dishes), inter-class similarity, number of images, number of classes, images acquisition condition, among others. 
 
\begin{figure}
\centering
\includegraphics[width=1\textwidth]{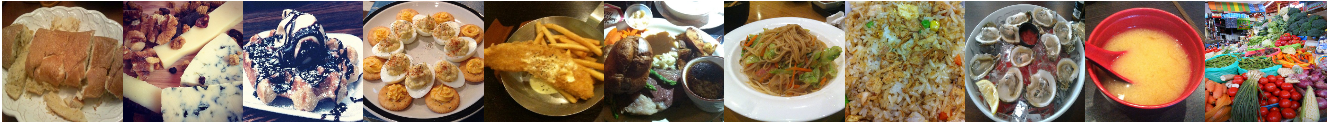}\caption{\small Images from the Food-11 dataset. Each image corresponds to a different class.}
\label{fig:fig2}
\end{figure}

\textbf{Food-11} is a dataset for food recognition \cite{singla2016}, which contains 16,643 images grouped into 11 general categories of food: pasta, meat, dessert, soup, etc. (see Fig. \ref{fig:fig2}).
The images were collected from existing food datasets (Food-101, UECFOOD100, UECFOOD256) and social networks (Flickr, Instagram). This dataset has an unbalanced number of images for each class with an average of 1,513 images per class and a standard deviation of 702.
For our experiments, we used the same data split, images and proportions, provided by the authors \cite{singla2016}. These are divided as 60\% for training, 20\% for validation and 20\% for test. In total, it consists of 9,866, 3,430 and 3,347 images for each set, respectively.  

\vspace*{-3mm}

\textbf{Food-101} is a standard to evaluate the performance of visual food recognition \cite{bossard2014}. This dataset contains 101.000 real-world food images downloaded from foodspotting.com, which were taken under unconstrained conditions. The authors chose the top 101 most popular classes of food (see Fig. \ref{fig:fig3}) and collected 1,000 images for each class: 75\% for training and 25\% for testing. With respect to the classes, these consist of very diverse and fine-grained dishes of various countries, but also with highly intra-class variation and inter-class similarity in most occasions. In our experiments, we used the same data splits provided by the authors. Unlike Food-11, and keeping the procedure followed by other authors \cite{liu2016,martinel2016,hassannejad2016}, in this dataset we directly validate and test on the same data split.

\subsection{Experimental Setup}

Every CNN model was previously trained on the ILSVRC dataset. Following, we adapted them by changing the output of the models to the number of classes for each target dataset and fine-tuned the models using the new images. For the training of the CNN models, we used the  Deep Learning framework Keras\footnote{www.keras.io}. 
The models chosen for Food-101 dataset due to their performance-efficiency ratio were InceptionV3 \cite{szegedy2016} and ResNet50 \cite{he2016}. Both models were trained during 48 epochs with a batch size of 32, and a learning rate of $5 \times 10^{-3}$ and $1 \times 10^{-3}$, respectively. In addition, we applied a decay of 0.1 during the training of InceptionV3 and of 0.8 for ResNet50 every 8 epochs. 
The parameters were chosen empirically by analyzing the training loss.

As to the Food-11 dataset, we kept the ResNet50 model, but changed InceptionV3 by GoogLeNet \cite{szegedy2015}, since InceptionV3 did not generalize well over Food-11. We believe that the reason is the small  number of images for each class not sufficient to avoid over-fitting; the model quickly obtained a good result in the training set, but a poor performance on the validation set. GoogLeNet and Resnet50 were trained during 32 epochs with a batch size of 32 and 16, respectively. The other parameters used for the ResNet50 were the same used for Food-101. In the case of GoogLeNet, we used a learning rate of $1 \times 10^{-3}$ and applied a decay of 0.1 during every 8 epochs, that turned out empirically the optimal parameters for our problem. 

\subsection{Data Preprocessing and Metrics}
The preprocessing made during the training, validation and testing phases was the following. 
During the training of our CNN models, we applied different preprocessing techniques on the images with the aim of increasing the samples and to prevent the over-fitting of the networks. First, we resized the images keeping the original aspect ratio as well as satisfying the following criteria: the smallest side of the resulting images should be greater than or equal to the input size of the model; and the biggest side should be less than or equal to the maximal size defined in each model to make random crops on the image. 
In the case of InceptionV3, we set to 320 pixels as maximal size,  for GoogLeNet and ResNet50 the maximal size was defined as 256 pixels. After resizing the images, inspired by \cite{hassannejad2016}, we enhanced them by means of a series of random distortions such as: adjusting color balance, contrast, brightness and sharpness. Finally, we made random crops of the images, with a dimension of 299x299 for InceptionV3 and of 224x224 for the other models. Then, we applied random horizontal flips with a probability of 50\%, and subtracted the average image value of the ImageNet dataset.
During validation, we applied a similar preprocessing, with the difference that we made a center crop instead of random crops and that we did not apply random horizontal flips. During test, we followed the same procedure than in validation (1-Crop evaluation). Furthermore, we also evaluated the CNN using 10-Crops, which are: upper left, upper right, lower left, lower right and center crop, both in their original setup and also applying an horizontal flip \cite{krizhevsky2012}. As for 10-Crops evaluation, the classifier gets a tentative label for each crop, and then majority voting is used over all predictions. In the cases where two labels are predicted the same number of times, the final label is assigned comparing their highest average prediction probability. 

We used four metrics to evaluate the performance of our approach, overall Accuracy (ACC), Precision (P), Recall (R), and $F_{1}$ score.

\subsection{Experimental Results on Food-11}

The results obtained during the experimentation on Food-11 dataset are shown in Table \ref{tab:tab1} giving the error rate (1 - accuracy) for the best CNN models, compared to the CNNs Fusion. We report the overall accuracy by processing the test data using two procedures: 1) a center crop (1-Crop), and 2) using 10 different crops of the image (10-Crops). 
The experimental results show an error rate of less than 10 \% for all  classifiers, achieving a slightly better performance when using 10-Crops. The best accuracy is achieved with our CNNs Fusion approach, which is about 0.75\% better than the best result of the classifiers evaluated separately. 
On the other hand, the baseline classification on Food-11 was given by their authors, who obtained an overall accuracy of 83.5\% using GoogLeNet models fine-tuned in the last six layers without any pre-processing and post-processing steps. Note that the best results obtained with our approach have been using the pointwise measures (S2, I2). The particularity of these measures is that they penalize big differences between corresponding values of DTs and DP being from the specific class to be assigned as the rest of the class values.
From now on, in this section we only report the results based on the 10-Crops procedure. 
 
\begin{table}
\vspace*{-2mm}
\caption{Overall test set error rate of Food-11 obtained for each model. The distance measure is shown between parenthesis in the CNNs Fusion models.} 
\vspace*{-3mm}
\centering
\setlength{\tabcolsep}{0.5em}
\begin{tabular}{c l | c c c}
	
    \textbf{Authors} &\multicolumn{1}{c}{\textbf{Model}} & \textbf{1-Crop} & \textbf{10-Crops} & \textbf{N/A} \\ 
    \hline 
    \cite{singla2016}&GoogLeNet & - & -& 16.5\% \\ 
    \hline
    \hline
    us &GoogLeNet & 9.89\% & 9.29\%& -\\ 
    us &ResNet50 & 6.57\% & 6.39\%& -\\ 
    \hline
    us &CNNs Fusion (S$_{1}$) & 6.36\% & 5.86\%& -\\ 
    us &CNNs Fusion (S$_{2}$) & 6.12\% & {\bf 5.65}\%& -\\ 
    us &CNNs Fusion (I$_{1}$) & 6.36\% & 5.89\%& -\\ 
    us &CNNs Fusion (I$_{2}$) & 6.30\% & {\bf 5.65}\%& -\\ 
    us &CNNs Fusion (C) & 6.45\% & 6.07\%& -\\ 
    us &CNNs Fusion (N) & 6.36\% & 5.92\%& -\\ 
    \hline
\end{tabular}
\label{tab:tab1}
\end{table}
\vspace*{-2mm}

As shown in Table \ref{tab:tab2}, the CNNs Fusion is able to properly classify not only the images that were correctly classified by both baselines, but in some occasions also when one or both fail. This suggests that in some cases both classifiers may be close to predicting the correct class and combining their outputs can make a better decision. 

\begin{table} 
\vspace*{-2mm}
\caption{Percentage of images well-classified and misclassified on Food-11 using our CNNs Fusion approach, distributed by the results obtained with GoogLeNet (CNN$_{1}$) and ResNet50 (CNN$_{2}$) models independently evaluated.}
\vspace*{-3mm} 
\centering
\setlength{\tabcolsep}{0.5em}
\begin{tabular}{l r r r r}
     & \multicolumn{4}{|c}{\textbf{CNNs evaluated independently}}\\ 
    \cline{2-5}
    \multicolumn{1}{c|}{\textbf{CNNs Fusion}} &
    \multicolumn{1}{c}{\textbf{Both wrong}} &
    \multicolumn{1}{c}{\textbf{CNN$_{1}$ wrong}} &
    \multicolumn{1}{c}{\textbf{CNN$_{2}$ wrong}} &
    \multicolumn{1}{c}{\textbf{Both fine}} \\
    \hline
    \hline
    \multicolumn{1}{l|}{\textbf{Well-classified}} & 3.08\% & 81.77\% & 54.76\% & 99.97\% \\ 
    \multicolumn{1}{l|}{\textbf{Misclassified}} & 96.92\% & 18.23\% & 45.24\% & 0.03\% \\ 
    \hline
\end{tabular}
\label{tab:tab2}
\end{table}
\vspace*{-2mm}

\begin{figure}
\centering
\includegraphics[width=1\textwidth]{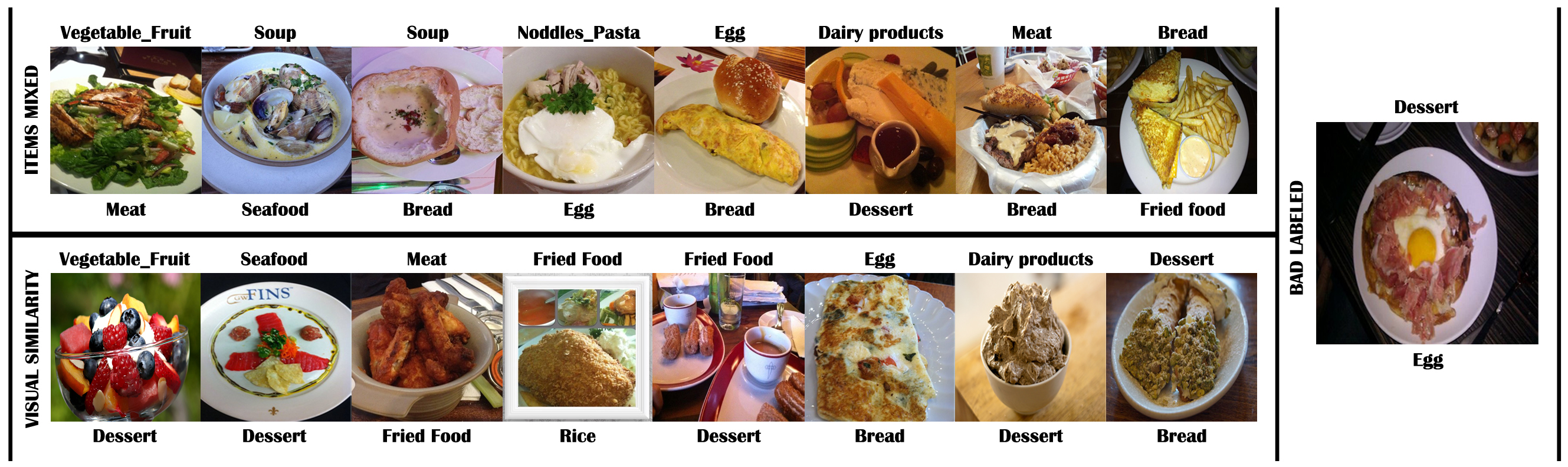}
\caption{\small Misclassified Food-11 examples: predicted labels (on the top), and the  groundtruth (on the bottom).}
\label{fig:fig4}
\end{figure}

Samples misclassified by our model are shown in Fig. \ref{fig:fig4},
where most of them are produced by mixed items, high inter-class similarity and wrongly labeled images. We show the ground truth (top) and the predicted class (bottom) for each sample image. 

In Table \ref{tab:tab3}, we show the precision, recall and $F_1$ score obtained for each class separately. By comparing the $F_1$ score, the best performance is achieved for the class Noodles\_Pasta and the worst for Dairy products. Specifically, the class Noddles\_Pasta only has one image misclassified, which furthermore is a hard sample, because it contains two classes together (see items mixed in Figure \ref{fig:fig4}). Considering the precision, the worst results are obtained for the class Bread, which is understandable considering that bread can sometimes be present in other classes. In the case of recall, the worst results are obtained for Dairy products, where an error greater than 8\% is produced for misclassifying several images as class Dessert. The cause of this is mainly, because the class Dessert has a lot of items in their images that could also belong to the class Dairy products (e.g. frozen yogurt or ice cream) or that are visually similar. 

\vspace*{-2mm}
\begin{table} \caption{Some results obtained on the Food-11 using our CNNs Fusion approach.} 
\vspace*{-3mm}
\centering
\setlength{\tabcolsep}{0.5em}
\begin{tabular}{l | r c c c}
	\hline
    \multicolumn{1}{c}{\textbf{Class}} & \multicolumn{1}{c}{\textbf{\#Images}} & \textbf{Precision} & \textbf{Recall} & \textbf{F1} \\ 
    \hline
    \hline
    Bread & 368 & \textbf{88.95\%} & 91.85\% & 90.37\% \\ 
    \hline
    Dairy products & 148 & 89.86\% & \textbf{83.78\%} & \textbf{86.71\%} \\ 
    \hline
    Meat & 432 & 94.12\% & 92.59\% & 93.35\% \\ 
    \hline
    Noodles\_Pasta & 147 & \textbf{100.00\%} & \textbf{99.32\%} & \textbf{99.66\%} \\ 
    \hline
    Rice & 96 & 94.95\% & 97.92\% & 96.41\% \\ 
    \hline
    Vegetable\_Fruit & 231 & 98.22\% & 95.67\% & 96.93\% \\ 
    \hline
\end{tabular}
\label{tab:tab3}
\end{table}

\vspace*{-.8cm}
\subsection{Experimental Results on Food-101}

The overall accuracy on Food-101 dataset is shown in Table \ref{tab:tab4} for two classifiers based on CNN models, and also for our CNNs Fusion. The overall accuracy is obtained by means of the evaluation of the prediction using 1-Crop and 10-Crops. The experimental results show better performance (about 1\% more) using 10-Crops instead of 1-Crop. From now on, in this section we only report the results based on the 10-Crops procedure. In the same way as observed in Food-11, the best accuracy obtained with our approach was by means of point-wise measures S2, I2, where the latter provides a slightly better performance.
Again, the best accuracy is also achieved by the CNNs Fusion, which is about 1.5\% higher than the best result of the classifiers evaluated separately. 
Note that the best performance on Food-101 (overall accuracy of 90.27\%) was obtained using WISeR \cite{martinel2016}.
In addition, the authors show the performance by another deep learning-based approaches, in which three CNN models achieved over a 88\% (InceptionV3, ResNet200 and WRN \cite{zagoruyko2016}). However, WISeR, WRN and ResNet200 models were not considered in our experiments since they need a multi-GPU server to replicate their results. In addition, those models have 2.5 times more parameters than the models chosen, which involve a high cost computational especially during the learning stage. Following the article steps, our best results replicating the methods were those using InceptionV3 and ResNet50 models used as a base to evaluate the performance of our CNNs Fusion approach. 

\vspace*{-2mm}
\begin{table} \caption{Overall test set accuracy of Food-101 obtained for each model.} 
\vspace*{-3mm}
\centering
\setlength{\tabcolsep}{0.5em}
\begin{tabular}{c l | c c c}
	
    \textbf{Author} &\multicolumn{1}{c}{\textbf{Model}} & \textbf{1-Crop} & \textbf{10-Crops} & \textbf{N/A} \\ 
    \hline
    \cite{hassannejad2016}&InceptionV3 & - & - & 88.28\% \\ 
    \cite{martinel2016}&ResNet200 & - & 88.38\%& - \\ 
    \cite{martinel2016}&WRN & - & 88.72\%& - \\ 
    \cite{martinel2016}&WISeR & - & 90.27\%& - \\ 
    \hline
    \hline
    us &ResNet50 & 82.31\% & 83.54\%& - \\ 
    us &InceptionV3 & 83.82\% & 84.98\%& - \\ 
    \hline
    us &CNNs Fusion (S$_{1}$) & 85.52\% & 86.51\%& - \\ 
    us &CNNs Fusion (S$_{2}$) & 86.07\% & 86.70\%& - \\ 
    us &CNNs Fusion (I$_{1}$) & 85.52\% & 86.51\%& - \\ 
    us &CNNs Fusion (I$_{2}$)& 85.98\% & {\bf 86.71}\%& - \\ 
    us &CNNs Fusion (C)& 85.24\% & 86.09\%& - \\ 
    us &CNNs Fusion (N)& 85.53\% & 86.50\%& - \\
    \hline

\end{tabular}
\label{tab:tab4}
\end{table}

\vspace*{-2mm}
As shown in Table \ref{tab:tab5}, in this dataset the CNNs Fusion is also able to properly classify not only the images that were correctly classified for both classifiers, but also when one or both fail. Therefore, we demonstrate that our proposed approach maintains its behavior independently of the target dataset.

\vspace*{-2mm}
\begin{table} \caption{Percentage of images well-classified and misclassified on Food-101 using our CNNs Fusion approach, distributed by the results obtained with InceptionV3 (CNN$_{1}$) and ResNet50 (CNN$_{2}$) models independently evaluated.} 
\vspace*{-3mm}
\centering
\setlength{\tabcolsep}{0.5em}
\begin{tabular}{l r r r r}
     & \multicolumn{4}{|c}{\textbf{CNNs evaluated independently}}\\ 
    \cline{2-5}
    \multicolumn{1}{c|}{\textbf{CNNs Fusion}} &
    \multicolumn{1}{c}{\textbf{Both wrong}} &
    \multicolumn{1}{c}{\textbf{CNN$_{1}$ wrong}} &
    \multicolumn{1}{c}{\textbf{CNN$_{2}$ wrong}} &
    \multicolumn{1}{c}{\textbf{Both fine}} \\
    \hline
    \hline
    \multicolumn{1}{l|}{\textbf{Well-classified}} & 1.95\% & 73.07\% & 64.95\% & 99.97\% \\ 
    \multicolumn{1}{l|}{\textbf{Misclassified}} & 98.05\% & 26.93\% & 35.05\% & 0.03\% \\ 
    \hline
\end{tabular}
\label{tab:tab5}
\end{table}

\vspace*{-2mm}
\begin{figure}
\centering
\includegraphics[width=1\textwidth]{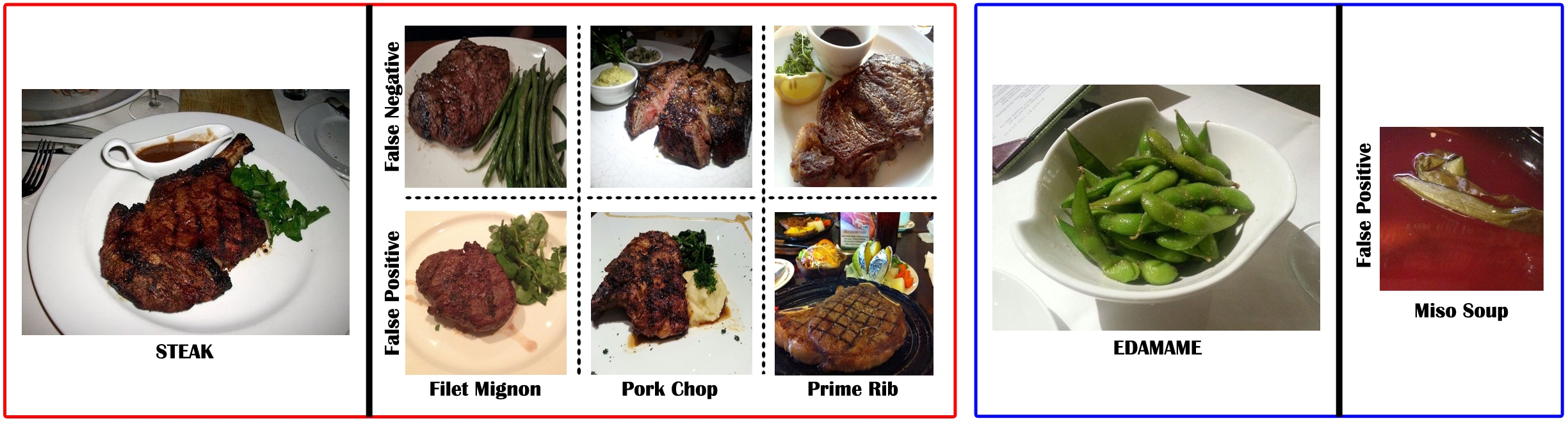}
\vspace*{-2mm}
\caption{\small Misclassified examples for the Food-101 classes that obtained the worst (steak) and best (edamame) classification results by F1 score (groundtruth label - bottom).}
\label{fig:fig5}
\end{figure}

Some examples of FPs and FNs are shown in Figure \ref{fig:fig5}. Analyzing the misclassified images like Steak, we chose a representive images per class corresponding to the top three classes with highest error, and in Edamame the unique image misclassified. Regarding the statistical results, in Table \ref{tab:tab6} are shown the top 3 better  and worst classification results on Food-101. We highlight the classes with the worst and best results. As for the worst class (Steak), the precision and recall achieved are 60.32\% and 59.60\%, respectively. Interestingly, about 26\% error in the precision and 30\% error in the recall is produced with only three classes: Filet mignon, Pork chop and Prime rib. As shown in Fig. \ref{fig:fig5}, these are fine-grained classes with high inter-class similarities that imply high difficulty for the classifier, because it is necessary to identify small details that allow to determine the corresponding class of the images. On the other hand, the best class (Edamame) was classified achieving 99.60\% of precision and 100\% of recall. Unlike Steak, Edamame is a simple class to classify, because it has a low intra-class variation and low inter-class similarities. In other words, the images in this class have a similar visual appearance and they are quite different from the images of the other classes. 

\vspace*{-4mm}
\begin{table} \caption{Top 3 better and worst classification results on Food-101.} 
\vspace*{-3mm}
\centering
\setlength{\tabcolsep}{0.5em}
\begin{tabular}{l | c c c}
	\hline
    \multicolumn{1}{c}{\textbf{Class}}  & \textbf{Precision} & \textbf{Recall} & \textbf{F1} \\ 
    \hline
    \hline
    Spaghetti Bolognese  & 94.47\% & 95.60\% & 95.03\% \\ 
    \hline
    Macarons  & 97.15\% & 95.60\% & 96.37\% \\ 
    \hline
    Edamame  & \textbf{99.60\%} & \textbf{100.00\%} & \textbf{99.80\%} \\ 
    \hline
    \hline
	Steak &  \textbf{60.32\%} & \textbf{59.60\%} & \textbf{59.96}\% \\ 
    \hline
    Pork Chop & 75.71\% & 63.60\% & 69.13\% \\ 
    \hline
    Foie Gras  & 72.96\% & 68.00\% & 70.39\% \\ 
  \hline
    \hline
\end{tabular}
\label{tab:tab6}
\end{table} 
\vspace*{-0.9cm}
\section{Conclusions}
\vspace*{-0.2cm}
\label{sec:conclusions}
In this paper, we addressed the problem of food recognition and proposed a CNNs Fusion approach based on the concepts of decision templates and decision profiles and their similarity that improves the classification performance with respect to using CNN models separately. Evaluating different similarity measures, we show that the optimal one is based on the infinimum of the maximum between the complementary of the decision templates and the decision profile of the test images. On Food-11, our approach outperforms the baseline 
accuracy by more than 10\% of accuracy. 
As for Food-101, we used two CNN architectures providing the best state of the art results where our CNNs Fusion strategy outperformed them again. 
As a future work, we plan to evaluate the performance of the CNN Fusion strategy as a function of the number of  CNN models. 
\vspace*{-0.2cm}
\section*{Acknowledgement}
\vspace*{-0.2cm}
This work was partially funded by TIN2015-66951-C2, SGR 1219, CERCA, \textit{ICREA Academia'2014}, CONICYT Becas Chile, FPU15/01347 and Grant 20141510 (Marat\'{o} TV3). We acknowledge Nvidia Corp. for the donation of Titan X GPUs.

\end{document}